\documentclass{Interspeech2024}

\interspeechcameraready

\usepackage[utf8]{inputenc} 
\usepackage[T1]{fontenc}    
\usepackage{url}            
\usepackage{booktabs}       
\usepackage{amsfonts}       
\usepackage{nicefrac}       
\usepackage{microtype}      
\usepackage{xcolor}         

\usepackage{amsmath}
\usepackage{mathtools}
\usepackage{graphicx}
\usepackage{caption}
\usepackage{subcaption}
\usepackage{pgfplots}
\usepackage{url}
\usepackage{tikz}

\definecolor{rebuttal}{HTML}{000000}

\newcommand{\Reals}{\mathbb R}

\newcommand{\Output}{\mathbf H}
\newcommand{\OutputVector}{\mathbf h}
\newcommand{\Input}{\mathbf X}
\newcommand{\InputVector}{\mathbf x}

\newcommand{\SummaryMean}{\bar{\mathbf s}}
\newcommand{\SummarySum}{\mathbf S}

\newcommand{\Transformation}{f}

\newcommand{\Summary}{s}
\newcommand{\NonLinearity}{\sigma}
\newcommand{\CombineFunction}{c}

\newcommand{\Time}{t}
\newcommand{\Length}{T}
\newcommand{\InputDimensionality}{D}
\newcommand{\OutputDimensionality}{D'}
\newcommand{\TransformedDimensionality}{D''}
\newcommand{\SummarySize}{D'''}

\newcommand{\Complexity}{\mathcal{O}}

\title{SummaryMixing: A Linear-Complexity Alternative to Self-Attention for Speech Recognition and Understanding}

\name[affiliation={}]{Titouan}{Parcollet$^*$}
\name[affiliation={}]{Rogier}{van Dalen$^*$}
\name[affiliation={}]{Shucong}{Zhang$^*$}
\name[affiliation={}]{Sourav}{Bhattacharya}

\address{
    Samsung AI Center Cambridge, United Kingdom}
\email{\{t.parcollet|r.vandalen|s1.zhang|sourav.b1\}@samsung.com}


\begin{document}

\maketitle
\def\thefootnote{*}\footnotetext{Equal Contribution}\def\thefootnote{\arabic{footnote}}

\begin{abstract}
%
Modern speech processing systems rely on self-attention.
Unfortunately, token mixing with self-attention takes quadratic time in the length of the speech utterance, slowing down inference and training and increasing memory consumption.
Cheaper alternatives to self-attention for ASR have been developed, but they fail to consistently reach the same level of accuracy.
This paper, therefore, proposes a novel linear-time alternative to self-attention.
It summarises an utterance with the mean over vectors for all time steps.
This single summary is then combined with time-specific information.
We call this method ``SummaryMixing''.
Introducing SummaryMixing in state-of-the-art ASR models makes it feasible to preserve or exceed previous speech recognition performance while making training and inference up to 28\,\% faster and reducing memory use by half.

\end{abstract}
\section{Introduction}

Automatic speech recognition (ASR) has greatly benefitted from deep learning \cite{nassif2019speech, arasteh2016iot}. However, in a push to improve recognition accuracy, ASR systems have steadily increased in model size.
Modern industry-scale ASR models often contain hundreds of millions or even billions of neural parameters \cite{radford2022robust}.
Training these models requires many GPU hours and results in large carbon footprints \cite{parcollet21_interspeech}.
Thus, this paper focuses on improving the efficiency of speech processing models.



At the core of current state-of-the-art (SOTA) speech systems are multi-head self-attention (MHSA) cells \cite{vaswani2017attention}. MHSA learns interactions between pairs of frames originating from the speech signal, and the interaction is also referred to as token mixing. 
Most state-of-the-art speech models use MHSA \cite{gulati2020conformer,peng2022branchformer}.
However, considering each pair of frames takes quadratic time in the input sequence length, making MHSA costly.

Recent works have pointed out that under some conditions, pair-wise self-attention operations in practice behave like linear operations. 
For example, \cite{zhang2021usefulness} first showed that the upper encoder layers in trained Transformer-based ASR models behave like feed-forward layers, which is also verified by \cite{shim2022understanding} for Conformer models. 
Furthermore, \cite{peng2022branchformer} demonstrated that the attention weights of trained Branchformer models tend to all have the same value, reducing the attention mechanism to computing an average.

Therefore, this work introduces an alternative to self-attention that takes only linear time in the sequence length.
Instead of computing pair-wise interactions, it summarises a whole utterance as a mean over a contribution for each time step. The obtained summary is then fed back to each time step. We call this method ``SummaryMixing''.

Our proposed SummaryMixing\footnote{\url{https://github.com/SamsungLabs/SummaryMixing}.} achieves a training time reduction of up to 28\% compared to MHSA. In decoding, its real-time factor does not increase with utterance length. SummaryMixing also halves the memory consumption in training and decoding, and reaches the performance of SOTA ASR systems on five datasets of different languages and acoustic conditions (Section \ref{sec:exps}). These findings extend to other speech understanding tasks including spoken language understanding (SLU) and keyword spotting (KWS). To the best of our knowledge, it is the first time a linear-time method matches or surpasses the performance of MHSA for speech-processing tasks across various scenarios.

\subsection{Related work}
\label{subsec:related work}

Numerous efficient attention mechanisms attempt to re-create the original behavior of self-attention but at a lower training cost.

Active research directions include low-rank approximation \cite{tay2022efficient}, linearization \cite{wang2020self}, or sparsification \cite{child2019generating} of self-attention.
In the context of ASR, the Squeezeformer \cite{kim2022squeezeformer}, the Efficient Conformer \cite{burchi2021efficient} and the Emformer \cite{shi2021emformer} reduce the length of the sequence attended to.
They lower training times and memory use by a constant factor, but retain the quadratic time complexity.

Some methods do provide linear complexity \cite{beltagy2020longformer,zaheer2020big,wang2020self,wu2021fastformer,samarakoon2022conformer}.
Fastformer \cite{wu2021fastformer}, the most successful linear alternative to self-attention will be used as a baseline for this paper.
On natural language processing tasks, its performance can be superior to MHSA.
Applied to speech recognition \cite{peng2022branchformer}, it achieved faster training yet slightly worse ASR performance than MHSA.

The other alternative from the literature is ContextNet \cite{han2020contextnet}, which is an entirely convolutional ASR system that reaches SOTA performance (though training times are not significantly improved).
Unfortunately, there exists no open-source implementation that reproduces the reported ASR results.

A recently proposed method called the HyperMixer \cite{mai2022hypermixer} derives from the MLP Mixer \cite{tolstikhin-2021-mlp}.
The MLP Mixer \cite{tolstikhin-2021-mlp} was the first to show that token mixing can also be achieved outside the framework of self-attention. 
MLP Mixer learns a fixed-size MLP to perform token mixing throughout time and achieves competitive performance across a wide range of domains with linear time complexity. 
The HyperMixer \cite{mai2022hypermixer} extends the MLP Mixer to variable-length sequences, while keeping the time complexity linear. Section \ref{subsec:summarymixing} will discuss this in more detail.

A longer version of this paper contains further detail%
\footnote{https://arxiv.org/pdf/2307.07421v2}.




\section{SummaryMixing}

Previous works \cite{zhang2021usefulness,peng2022branchformer} have shown that ASR does not require long-distance fine-grained modeling at the acoustic level. This section introduces SummaryMixing (section \ref{subsec:summarymixing})  and its integration into the Branchformer and Conformer architectures (section \ref{subsec:branchformer}).

\subsection{SummaryMixing}
\label{subsec:summarymixing}

\begin{figure}[!t]
    \begin{subfigure}[b]{.50\columnwidth}
        \centering
        \includegraphics{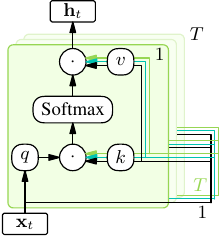}
        \label{figure:self-attention}
    \end{subfigure}
    \hfill
    \begin{subfigure}[b]{.38\columnwidth}
        \centering
        \includegraphics{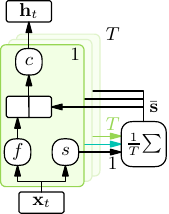}
        \label{figure:summary_mixing}
    \end{subfigure}
    \\
    \caption{ Comparison of the self-attention cell (left) and the newly proposed SummaryMixing cell (right). In SummaryMixing, the information from all time steps is averaged, and this average is fed back to each time step $T$.}
    \label{fig:architectures}
\end{figure}

Figure \ref{fig:architectures} shows a self-attention cell \cite{vaswani-2017-attention}.

The plate and its content are replicated for each time step $\Time = 1 \ldots \Length$.
The cell takes an input sequence $\Input \in \Reals^{\Length \times \InputDimensionality} = \{\InputVector_1,\ldots,\InputVector_{\Length}\}$ of $\Length$ feature vectors $\InputVector_{\Time}$ of length $\InputDimensionality$, and transforms them into hidden representations $\Output \in \Reals^{\Length \times \OutputDimensionality} = \{\OutputVector_1,\ldots,\OutputVector_{\Length}\}$, that can be inputs for the next layer.
The self-attention cell computes a weighted average, where the weights are computed for each time step, of ``values'' (computed with $v$) for each time step.
The connections across time steps in the left part of Figure \ref{fig:architectures} highlight the quadratic cost in the length of the input that MHSA takes.

To reduce the quadratic time complexity, we introduce SummaryMixing, which also transforms input vectors $\InputVector_{\Time}$ into hidden representations $\OutputVector_{\Time}$.
The key to inducing linear time complexity is to summarise the whole utterance in a single vector $\SummaryMean$.
Figure \ref{fig:architectures} illustrates this.
The input $\InputVector_{\Time}$ is transformed by two functions.
One is the local transformation function $\Transformation: \Reals^{\InputDimensionality} \rightarrow \Reals^{\TransformedDimensionality}$.
The other is summary function $\Summary: \Reals^{\InputDimensionality} \rightarrow \Reals^{\SummarySize}$.
The resulting vectors $\Summary(\InputVector_{\Time})$ are averaged across all time steps ($\frac1{\Length}\!\sum$) to form the mean vector $\SummaryMean$.
This single vector is passed back to each time step.
The concatenation of it and the local information $\Transformation(\InputVector_{\Time})$ is then transformed by the combiner function $\CombineFunction: \Reals^{\TransformedDimensionality + \SummarySize} \rightarrow \Reals^{\OutputDimensionality}$. The SummaryMixing process can be described as:
\begin{align}
    \SummaryMean &= \frac1{\Length} \sum_{\Time=1}^{\Length} \Summary(\InputVector_{\Time})
    ;&
    \OutputVector_{\Time} &= \CombineFunction(\Transformation(\InputVector_{\Time}), \SummaryMean).
    \label{eq:summarymixing}
\end{align}%
Each output vector is the function of one vector capturing the whole sequence and one capturing local information.
Computing $\SummaryMean$ takes $\Complexity(\Length)$ time, after which each $\OutputVector_{\Time}$ can be computed in constant time w.r.t.\ $\Length$.
This compares to $\Complexity(\Length^2)$ in MHSA.

\textbf{Relationship to the HyperMixer.}
The HyperMixer was proposed by \cite{mai2022hypermixer} as a more efficient alternative for self-attention.
Though \cite{mai2022hypermixer} fails to mention this, the HyperMixer, like SummaryMixing, takes linear time in the sequence length.
Disecting the relationship is tortuous, so the full analysis is relegated to the appendix of an extended version of this paper%
\footnote{\url{https://arxiv.org/pdf/2307.07421v2}}.
In brief, the HyperMixer turns out to have the form
\begin{align}
    \SummarySum &= \textstyle
        \NonLinearity\big(\sum_{\Time=1}^{\Length} \Transformation'(\InputVector_{\Time}) \times \InputVector_{\Time} \big) ;
    &
    \OutputVector_{\Time} &= \SummarySum \cdot \Transformation(\InputVector_{\Time})
    .
\end{align}
The key similarity to SummaryMixing is that the only interaction between any two feature vectors is through a sum over all $\Time$.

\subsection{{Branchformer and Conformer with SummaryMixing}}
\label{subsec:branchformer}

The Branchformer \cite{peng2022branchformer} and Conformer \cite{gulati2020conformer} reach state-of-the-art accuracy in speech recognition and understanding. Both architectures contain CNN and MHSA blocks responsible for capturing local and global dependencies respectively. We propose to replace MHSA with SummaryMixing.

In particular, the transformation ($\Transformation$), summary ($\Summary$), and combiner ($\CombineFunction$) functions are all implemented as a dense linear layer followed by a GeLU activation function. 
The input of the combiner is a concatenation of $\SummaryMean$ and $\Transformation(\InputVector_{\Time})$. 
The CNN branch of the Branchformer is an MLP with convolutional gating inspired by the cgMLP of \cite{sakuma2021mlp}. The outputs of both branches, CNN and SummaryMixing, are then concatenated and fed to a two-layered MLP followed by GeLU activations before feeding into the next block. For the Conformer, the output of the SummaryMixing block is simply fed to the next convolutional module.

\section{Experiments}
\label{sec:exps}

First, empirical efficiency gains in terms of training speed, memory consumption, as well as real-time decoding factors (RTF), are highlighted in controlled tasks (Section \ref{subsec:efficiency_analysis}). Then, the compared models are applied to standard ASR and SLU evaluations with common datasets and architectures (Section~\ref{subsec:exps_ctcatt}).  

\subsection{Experimental Protocol}
\label{subsec:exps_protocol}

Different hyperparameters, architectures, and datasets are described in the corresponding sections while the baselines and the framework are shared and described here. The exhaustive list of hyperparameters of every experiment can be found in the public code repository.

\textbf{Baseline architectures.} All ASR models are based on the encoder-decoder architecture \cite{karita2019comparative} with joint CTC/Transformer training, except for the efficiency and RTF analysis where the input vectors are random, so a simple CTC decoder is used instead to avoid any side effects. Classification tasks are performed by adding a classifier on top of the averaged encoder representation over time. Models are compared by varying the encoder architecture based on the literature.


\begin{figure*}[!t]
\begin{center}

\includegraphics{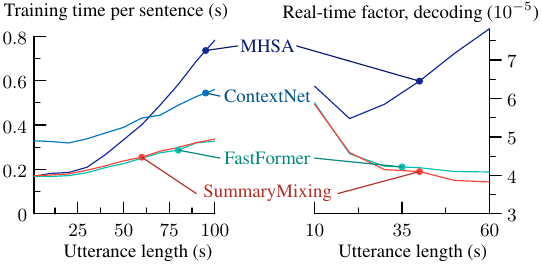}%
\hspace*{0.2cm}%
\includegraphics{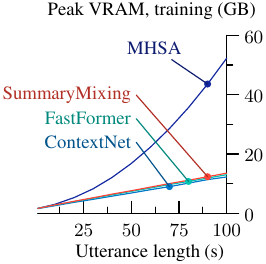}

\caption{Efficiency measurements and real-time factor analysis. The left- and right-most curves represent the average time as well as the peak VRAM consumption to process a sequence of various lengths. The curve in the middle shows the RTF for trained ASR systems. } \label{fig:curves}
\end{center}
\end{figure*}

We consider the following encoders as baselines.
First, two common architectures:
the Conformer \cite{gulati2020conformer}, which is currently employed in most deployed ASR systems in real-world products \cite{guo2021recent}; and the Branchformer, which is an improvement over the Conformer \cite{peng2022branchformer}.
Then, two baselines with linear-complexity alternatives to attention.
First, a Branchformer equipped with the FastFormer \cite{wu2021fastformer, peng2022branchformer}.
The Fastformer has outperformed the best linear alternatives to self-attention in various tasks \cite{wu2021fastformer}.
We also introduce a Branchformer equipped with HyperMixer attention \cite{mai2022hypermixer}.
The final two baselines use no attention, but only CNNs.
The first is ContextNet, first introduced by \cite{han2020contextnet}, which has shown competitive performance.
To serve as a low-bar baseline, a the MHSA branch is removed from a Branchformer, leaving only the convolutional branch.

\textbf{Implementation details.} ASR and SLU systems have been implemented within the SpeechBrain toolkit \cite{ravanelli2021speechbrain} version 0.5.15. Experiments are conducted following officially available and open-source recipes%
    \footnote{\url{https://github.com/SamsungLabs/SummaryMixing}}%
, changing only the architecture of the models to allow for fair comparison and easy replication. Reported results are obtained from training and not the literature.

\subsection{Efficiency and Real-Time Factor Analysis}
\label{subsec:efficiency_analysis}

First, controlled experiments are run to examine the growth in training time, decoding speed, and VRAM requirements as a function of the input utterance length.
These experiments use synthetic and manipulated data.

\textbf{Efficiency task details.} Systems are benchmarked both in terms of measured training time and peak VRAM consumption. In practice, five thousand sequences of random tensors corresponding to a signal of length $L$ with $1\le L \le100$ in seconds and sampled at 16 kHz are generated and used as inputs, while one hundred random tokens corresponding to indices in a vocabulary of size 1,000 are used as targets. The final training time is an average and is expressed in seconds. Measurements are extracted on an isolated compute node with four Tesla A100 80GB and bf16 mixed precision \cite{kalamkar2019study}.

\textbf{Real-time factor task details.} Real Time Factor (RTF) measurements are obtained by dividing the time taken to decode an utterance by its duration.
All models are trained on Librispeech (with WERs reported in Table~\ref{tab:res_libri}).
They are compared by varying the length of the utterances to decode.
We constructed six sets of 2,500 sentences of duration ${10,20,30,40,50,60}$ seconds by taking utterances from Librispeech \textit{test-clean} and either cropped them to the desired duration or concatenating multiple utterances to reach longer lengths (typically ${30,40,60}$ seconds).
Models are compared on the same sets of sentences. Batched greedy CTC decoding is applied.

\textbf{Model architectures.} The selected models for these experiments are the Branchformer equipped with MHSA, FastFormer, SummaryMixing as well as ContextNet.
The encoders have 18 layers with dimensionality 512; the decoders 6 layers with dimensionality 256.
The number of neural parameters is roughly 80M for the Branchformer with MHSA, and 65M for ContextNet.
This corresponds to architectures reaching state-of-the-art WER on Librispeech.
The Branchformer with SummaryMixing also has 80M parameters.
The architecture for ContextNet follows the architecture description described in the original work \cite{han2020contextnet} leading to 65M parameters.

\subsubsection{Results and discussion}

Figure \ref{fig:curves} depicts the obtained efficiency and RTF measurements for the considered models. It is clear from the training time, RTF as well as peak VRAM consumption that the MHSA-equipped Branchformer leads to a quadratic increase in required resources.
For instance, with speech utterances of duration 100 seconds, training takes 2.5 times longer than the SummaryMixing Branchformer and the VRAM consumption explodes from 11.6 GB for the SummaryMixing Branchformer to 52 GB for the MHSA Branchformer. The latter phenomenon is particularly critical as it drastically impacts the price and availability of the required hardware.
The contrast is even starker for the real-time factor of decoding.
For the linear-time attention mechanisms, the FastFormer and SummaryMixing, the real-time factor asymptotes as the utterances grow longer.
For MHSA, the real-time factor grows linearly with the length of the utterance.

\subsection{Speech Recognition and Understanding Experiments}
\label{subsec:exps_ctcatt}

This section examines SummaryMixing for ASR and SLU systems.

\textbf{Speech recognition tasks details.} ASR is conducted on five datasets of different languages and complexities in terms of acoustic conditions and available training data: LibriSpeech \cite{panayotov2015librispeech}, CommonVoice (version 13.0) \cite{ardila2020common} Italian (300 hours), Dutch (40 hours), and French (730 hours) as well as AISHELL-1 \cite{bu2017aishell} and Ted-Lium 2 \cite{rousseau2014enhancing}. Evaluations are conducted on the official sets of each dataset. On Librispeech, models are evaluated without an LM on the \textit{dev-clean} set, and compared with a transformer LM shallow fusion on the \textit{test-clean} and \textit{test-other}. No language models are used for the other datasets.

\textbf{Speech understanding tasks details.} Models are compared across two tasks of speech understanding with the SLURP dataset from \cite{bastianelli2020slurp} for scenarios, actions, and entity classification and the Google speech commands dataset for keyword spotting.

\textbf{Model architectures.} All baselines are selected for ASR with Librispeech as an initial performance and training cost analysis. Then, a reduced subset of the baselines including the Branchformer with MHSA, SummaryMixing, and FastFormer is used across the seven other datasets for further comparison. We stick to the original recipe of SpeechBrain to start from a state-of-the-art Branchformer and Conformer, leading to model sizes of roughly 110M parameters for all considered methods except the SummaryMixing Conformer (103M), the FastFormer Branchformer (101M), the ContextNet (100M), and the CNN-only Branchformer (77M).
Models are multi-task trained with CTC and a Transformer decoder.
The Transformer language model is pre-trained and obtained from the SpeechBrain toolkit. 

\subsubsection{Speech recognition analysis on Librispeech}
\label{subsubsec:libri}

\begin{table}[t!]
    \centering
    \caption{Speech recognition results on encoder-decoder models with CTC plus Transformer decoding on the Librispeech dataset. ``GPU hours'' is the total training time. ``VRAM'' reports the peak amount of VRAM over the four GPUs during training.}
    \label{tab:res_libri}
    \begin{tabular}{@{~}l@{~~}l@{}c@{~}c@{~}c@{~~}c@{~}c@{~}}
    \toprule
        \textbf{Encoder} & \textbf{Variant} & \multicolumn{3}{c}{\textbf{WER} \%} & \textbf{GPU} & \textbf{VRAM}  \\
        && \emph{dev-} & \multicolumn{2}{c}{\emph{test-}} & hours & GB  \\
        && \emph{\llap{c}lean} & \emph{clean} & \emph{othe\rlap{r}} \\
    \midrule
           ContextNet & --- & 3.3 & 2.3 & 5.9  & 160  & 25 \\
           Conformer & {Self-attention} & \textbf{2.8} & 2.3 & 5.4  & 137  & 46  \\
           Branchformer & {Self-attention} & 2.9 & 2.2 & \textbf{5.1}  & 132 &  45  \\
           & {CNN Only} & 3.1 & 2.4 & 5.7  & \textbf{83} & 22  \\
           & {HyperMixer} & 3.1 & 2.3 & 5.6  & 126 & 30  \\
           & {FastFormer} & 3.0 & 2.2 & 5.4  & 96 & 23 \\
           \midrule
           \textbf{Proposed} \\
           Conformer
           & SummaryMix. & \textbf{2.8} & \textbf{2.1} & \textbf{5.1}  & 98 &  \textbf{21}  \\
           Branchformer
           & SummaryMix. & 2.9 & 2.2 & \textbf{5.1}  & 105 &  26 \\
           \multicolumn{2}{@{~}l}{~ + SummaryMixing decoder}   & 3.1 & 2.3 & 5.3  & 104 &  26
           \\ 
    \bottomrule
    \end{tabular}
\end{table}

Table \ref{tab:res_libri} lists the word error rates (WERs) as well as the total training time and peak VRAM consumption on Librispeech. All models, including the CNN-only alternatives, achieve competitive recognition rates. For instance, the CNN-only Branchformer achieved a WER of 3.1\% on \textit{dev-clean}. This finding supports the evidence that MHSA may not be necessary for the encoder of speech recognizer systems to achieve good accuracies. It is interesting to notice, however, that using MHSA to incorporate the global context slightly improves the overall word error rates while strongly impacting the needed resources. In fact, the 0.2\%, 0.2\%, and 0.6\% improvements on the \textit{dev-clean}, \textit{test-clean} and \textit{test-other} sets respectively of the MHSA Branchformer compared to the CNN-only Branchformer is done at the expense of 49 hours of compute, representing an increase of 58\% in training time. The VRAM goes from 22 GB to 45 GB for the CNN-only and MHSA versions respectively (i.e.\ an increase of 105\%). 

SummaryMixing Branchformers and Conformers reduce this disproportionate resource impact while preserving or improving the performance. The SummaryMixing Branchformer closes the gap with MHSA by achieving strictly the same performance with a reduction of the peak VRAM from 45 GB to 26 GB. ContextNet, despite achieving respectable performance, is not at the level initially reported by \cite{han2020contextnet}. However, there exists no replication of such results. Finally, the SummaryMixing Conformer also beats the standard Conformer with MHSA and reaches the best \textit{test-clean} and \textit{test-other} WER among all the models while halving the required VRAM from 46 GB for the MHSA variant to 21 GB and exhibiting a 28\% faster training.

\textbf{Removing the attentional decoder.}
Apart from in encoders, in decoders attention is also used, both cross-attention and self-attention.
Replacing cross-attention in the decoder leads to severe performance degradation.
On the other hand, replacing just self-attention in the decoder with SummaryMixing leads to reasonable performance, as shown in the last row of Table \ref{tab:res_libri}.
However, the gains in efficiency are not as impressive as when replacing attention in the encoder, since the sequence length of the decoded text is much lower than of the input audio.

It is also interesting to remove the decoder entirely, so that the system uses no attention at all.
For this, we decode with CTC only (though the training loss is still CTC plus attention).
Without any LM, the Branchformer with MHSA obtains 2.6\% and 6.2\% of WER on \textit{test-clean} and \textit{test-other} sets, compared to 2.5\% and 6.4\% with SummaryMixing.
Such numbers are comparable to WeNet \cite{zhang2022wenet}. Finally, we also performed CTC-only training following the official SpeechBrain recipes on the two most promising architectures from Librispeech: Conformer with MHSA and SummaryMixing. With CTC greedy decoding (not in the table), on the \textit{dev-clean}, \textit{test-clean}, and \textit{test-other}, the Conformer with MHSA (28.8M) achieves WERs of 3.5\%, 3.7\%, 9.2\% respectively while the SummaryMixing-Conformer (26.5M) reaches WERs of 3.5\%, 3.7\%, and 9.4\%.

\begin{table}[t!]
    \centering
    \caption{Speech recognition, keyword spotting, and speech understanding results. ASR accuracy is expressed in word error rate. SLU results are expressed in SLU-F1 for SLURP and accuracy for Google Speech Command (GSC). }
    \label{tab:res_cv}
    \begin{tabular}{l@{~~}c@{~~}c@{~~}c@{~~}c@{~~}c@{~~}c@{~~}c@{~~}}
    \toprule 
    \textbf{Metric} &  \multicolumn{5}{c}{ASR WER $\downarrow$}
    & F1 $\uparrow$ & Acc. $\uparrow$ \\
        \textbf{Encoder} &  \textbf{Nl.} & \textbf{It.} & \textbf{Fr.} & \textbf{AI.} & \textbf{Ted.}
       & \textbf{SLURP} & \textbf{GSC}\\
       \hline
           Branchformer  & 32.6  & 10.5 & 11.0 & \textbf{5.7} & 7.9
           & 0.771 & 98.06 \\
           \textemdash \textit{FastFormer}   & 33.9   & 10.9 & 10.9 & 6.1 & 8.5 & --- & --- \\
           \textemdash \textit{Sum.Mix.}   & \textbf{31.5}   & \textbf{10.4} & \textbf{10.8} & \textbf{5.7} & \textbf{7.8}\
           & \textbf{0.773} & \textbf{98.16} \\
    \bottomrule
    \end{tabular}
\end{table}

\subsubsection{Extended ASR and SLU experiments}

We focused on the Branchformer with MHSA or FastFormer versus SummaryMixing Branchformer comparison by extending the number of languages and acoustic conditions for ASR. We also introduce two more tasks including keyword spotting and SLU. Results are reported in Table \ref{tab:res_cv}. 
Focusing on ASR, it is worth noting that the SummaryMixing Branchformers outperform the MHSA Branchformer in terms of WER with all datasets, and hence in all data regimes. Indeed, on average over all the model sizes and datasets, the SummaryMixing Branchformer reduces the WER by 0.5\% absolute compared to the MHSA Branchformer. This conclusion also extends to other tasks as the SummaryMixing Branchformers can reach their MHSA counterpart on spoken language understanding with an SLU-F1 score of 0.773 for the SummaryMixing compared to 0.771 for MHSA. For keyword spotting, SummaryMixing improves the accuracy over MHSA by 0.1\% absolute.

\section{Conclusion}

This work has proposed SummaryMixing, a novel linear-time complexity block removing the need for self-attention in speech recognition and understanding encoders. SummaryMixing is based on the following assumptions: (a) the acoustic modeling component of speech systems does not require multi-head self-attention; and (b) an efficient and cheaper global context vector taking the form of a summary of each speech utterance is sufficient to reach top-of-the-line speech recognition and understanding. SummaryMixing-equipped Conformers and Branchformers outperform state-of-the-art MHSA-equipped equivalent architectures while exhibiting a linear complexity, leading to a reduction of up to 28$\%$ in training time as well as more than half of the original VRAM consumption. SummaryMixing also leads to significantly faster inference and decoding times for offline speech recognition and understanding and could be extended to any speech encoder.

\newpage

\bibliographystyle{IEEEtran}
\bibliography{literature}


\end{document}